\PassOptionsToPackage{dvipsnames}{xcolor}
\documentclass[11pt,a4paper]{article}
\usepackage[hyperref]{emnlp-ijcnlp-2019}
\usepackage{times}
\usepackage{latexsym}
\usepackage{comment}
\usepackage{url}
\usepackage{times}
\usepackage{subcaption}
\usepackage{todonotes}
\usepackage{booktabs,tabularx, colortbl}
\usepackage{graphicx}
\usepackage{arydshln}
\usepackage{amsmath}
\usepackage{bm}
\usepackage[normalem]{ulem}
\usepackage[inline]{enumitem}
\usepackage[capitalize]{cleveref}
\usepackage{xspace}
\usepackage{soul}
\usepackage{titlesec}
\usepackage{blindtext}
\usepackage{scrextend}
\usepackage{amsmath}
\usepackage{tikz}
\newcommand*\circled[1]{\tikz[baseline=(char.base)]{
            \node[shape=circle,draw,inner sep=0.4pt] (char) {#1};}}

\addtokomafont{labelinglabel}{\sffamily}
\aclfinalcopy 

\crefformat{section}{\S#2#1#3} 
\crefformat{subsection}{\S#2#1#3}
\crefformat{subsubsection}{\S#2#1#3}
\crefmultiformat{section}{\S#2#1#3} { , \S#2#1#3}{, \S#2#1#3}{ and \S#2#1#3}
\crefrangeformat{section}{\S#3#1#4 to \S#5#2#6}

\definecolor{purple}{HTML}{9900ff}

\title{An Entity-Driven Framework for Abstractive Summarization}

\author{Eva Sharma$^{1}$\thanks{\quad These authors contributed equally. Work done while LH was at Northeastern University.} \quad Luyang Huang$^{2}$\footnotemark[1] \quad Zhe Hu$^{1}$\footnotemark[1] \quad {\rm and} \quad Lu Wang$^{1}$\\
  $^{1}$Khoury College of Computer Sciences, Northeastern University, Boston, MA 02115\\
  $^{2}$Department of Electrical and Computer Engineering, Boston University, Boston, MA 02215\\
  $^{1}${\tt evasharma@ccs.neu.edu, hu.zhe@husky.neu.edu, luwang@ccs.neu.edu} \\
  $^{2}${\tt lyhuang@bu.edu} \\
  }
\date{}

\mathcode`*=\string"8000
\begingroup
\catcode`*=\active
\xdef*{\noexpand\textup{\string*}}
\endgroup

\begin{document}
\maketitle

\begin{abstract}
%

Abstractive summarization systems aim to produce more coherent and concise summaries than their extractive counterparts. Popular neural models have achieved impressive results for single-document summarization, yet their outputs are often incoherent and unfaithful to the input. 
In this paper, we introduce \textbf{SENECA}, a novel System for ENtity-drivEn Coherent Abstractive summarization framework that leverages entity information to generate informative and coherent abstracts. 
Our framework takes a two-step approach: (1) an \textit{entity-aware content selection} module first identifies salient sentences from the input, then (2) an \textit{abstract generation} module conducts cross-sentence information compression and abstraction to generate the final summary, which is trained with rewards to promote coherence, conciseness, and clarity. 
The two components are further connected using reinforcement learning. 
Automatic evaluation shows that our model significantly outperforms previous state-of-the-art on ROUGE and our proposed coherence measures on New York Times and CNN/Daily Mail datasets. Human judges further rate our system summaries as more informative and coherent than those by popular summarization models.

\end{abstract}

\section{Introduction}
\label{sec:introduction}
Automatic abstractive summarization carries strong promise for producing concise and coherent summaries to facilitate quick information consumption~\cite{luhn1958automatic}. 
Recent progress in neural abstractive summarization has shown end-to-end trained models~\cite{nallapati2016abstractive,tan2017abstractive,celikyilmaz2018deep,kryscinski2018improving} excelling at producing fluent summaries.
Though encouraging, their outputs are frequently found to be unfaithful to the input and lack inter-sentence coherence~\cite{cao2017faithful,see-etal-2017-get,wiseman-etal-2017-challenges}. These observations suggest that existing methods have difficulty in identifying salient entities and related events in the article~\cite{fan2018robust}, and that existing model training objectives fail to guide the generation of coherent summaries.

\begin{figure}[t]
    \def\arraystretch{1.5}
    \bgroup
    \def\arraystretch{1.5}
	\fontsize{9}{10}\selectfont
     \hspace{-1mm}
	\setlength{\tabcolsep}{0.8mm}
	\begin{tabular}{|p{75mm}|}
	\hline
	\textbf{Input Article}: 
	
    $\ldots$ {\color{blue!75} \textbf{Prime Minister Bertie Ahern}} of Ireland called Sunday for a general election on May 24. 
    
    [{\color{blue!75} \textbf{Mr. Ahern}}] and his centrist party have governed in a coalition government since 1197 $\ldots$
    
    Under Irish law, which requires legislative elections every five years, \underline{{\color{blue!75} \textbf{Mr. Ahern}} had to call elections by midsummer.} 
    
    On Sunday, \{{\color{blue!75} \textbf{he}}\} said {\color{blue!75} \textbf{he}} would base {\color{blue!75} \textbf{his}} campaign for re-election on {\color{blue!75} \textbf{his}} work to strengthen the economy and efforts to revive {\color{red!70} \textbf{Northern Ireland's}} stalled peace process this year. 
    
    Political analysts said they expected {\color{blue!75} \textbf{Mr. Ahern 's}} work in {\color{red!70} \textbf{Northern Ireland}} to be an asset $\ldots$ \\
	
	\hline
	
	\textbf{Human Summary}: 
	
    {\footnotesize \circled{1}} {\color{blue!75} \textbf{Prime Min Bertie Ahern}} of Ireland calls for general election on May 24. 
    
    {\footnotesize \circled{2}} [\underline{{\color{blue!75} \textbf{He}}] is required by law to call elections by midsummer.}
    
    {\footnotesize \circled{3}} Opinion polls suggest {\color{blue!75} \textbf{his}} centrist government is in danger of losing its majority in Parliament because of public disgruntlement about overburdened public services. 
    
    {\footnotesize \circled{4}} \{{\color{blue!75} \textbf{Ahern}}\} says {\color{blue!75} \textbf{he}} would base {\color{blue!75} \textbf{his}} campaign for re-election on {\color{blue!75} \textbf{his}} work to strengthen economy and {\color{blue!75} \textbf{his}} efforts to revive {\color{red!70} \textbf{Northern Ireland's}} stalled peace process. 
    
    {\footnotesize \circled{5}} Analysts expect {\color{blue!75} \textbf{his}} work in {\color{red!70} \textbf{Northern Ireland}} to be asset.\\ 
    \hline

	\end{tabular}
    \vspace{-3mm}
\caption{Sample summary of an article from the New York Times corpus~\cite{sandhaus2008new}.
Mentions of the same entity are colored. \underline{Underlined} sentence in the article occurs relatively at an earlier position in the summary ({\footnotesize \circled{2}}) to improve topical coherence. Mentions in brackets (``[]",``\{\}") show different ways in which the same entity is referred to in the article and the summary.
Detailed explanation is given in \cref{sec:introduction}.
} 
\label{fig:sample-keyphrase}
\egroup
\vspace{-2.5mm}
\end{figure}
\begin{figure}[t]
    \centering
    \includegraphics[width=\columnwidth,trim=0.55cm 0 0.1cm 0, clip]{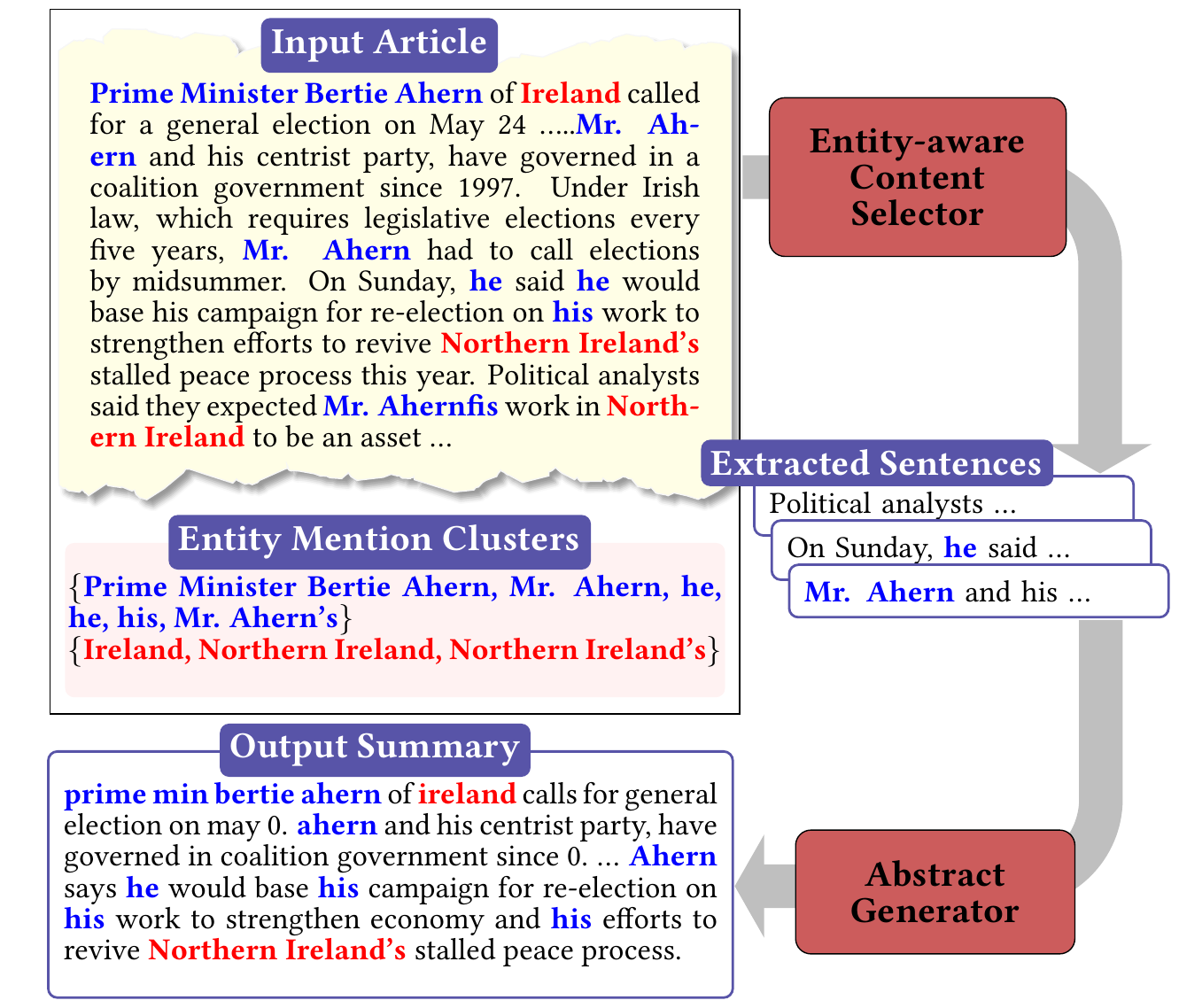}
    \vspace{-7mm}
    \captionof{figure}{ 
    Our proposed entity-driven abstractive summarization framework. Entity-aware content selector extracts salient sentences and abstract generator produces informative and coherent summaries. Both components are connected using reinforcement learning.  
    }
    \label{fig:pipeline}
    \vspace{-4mm}
\end{figure}
In this paper, we present \textbf{SENECA}, a {S}ystem for {EN}tity-driv{E}n {C}oherent {A}bstractive summarization.\footnote{Our code is available at \href{https://evasharma.github.io/SENECA}{evasharma.github.io/SENECA}.} We argue that entity-based modeling enables enhanced {\it input text interpretation}, {\it salient content selection}, and {\it coherent summary generation}, three major challenges that need to be addressed by single-document summarization systems~\cite{jones1999automatic}. We use a sample summary in \cref{fig:sample-keyphrase} to show entity usage in summarization. Firstly, frequently mentioned entities from the input, along with their contextual information, underscores the salient content of the article~\cite{nenkova2008entity}. 
Secondly, as also discussed in prior work~\cite{barzilay2008modeling,siddharthan2011information}, patterns of entity distributions and how they are referred to contribute to the coherence and conciseness of the text. For instance, a human writer places the underlined sentence in the input article next to the first sentence in the summary to improve topical coherence as they are about the same topic (``elections").  Moreover, the human often optimizes on conciseness by referring to entities with pronouns (e.g., ``he") or last names (e.g., ``Ahern") without losing clarity.

We therefore propose a two-step neural abstractive summarization framework to emulate the way humans construct summaries with the goal of improving both informativeness and coherence of the generated abstracts. 
As shown in \cref{fig:pipeline}, an {\bf entity-aware content selection} component first selects important sentences from the input that includes references to salient entities. 
An {\bf abstract generation} component then produces coherent summaries by conducting cross-sentence information ordering, compression, and revision. Our abstract generator is trained using reinforcement learning with rewards that promote informativeness and optionally boost coherence, conciseness, and clarity of the summary. 
To the best of our knowledge, we are the first to study coherent abstractive summarization with the inclusion of linguistically-informed rewards. 
%

We conduct both automatic and human evaluation on popular news summarization datasets. 
Experimental results show that our model yields significantly better ROUGE scores than previous state-of-the-art~\cite{gehrmann-etal-2018-bottom,celikyilmaz2018deep} as well as higher coherence scores on the New York Times and CNN/Daily Mail datasets. 
%
Human subjects also rate our system generated summaries as more informative and coherent than those of other popular summarization models. 

\section{Summarization Framework}
\label{sec:model}

In this section, we describe our entity-driven abstractive summarization framework which follows a two-step approach as shown in \cref{fig:pipeline}. It comprises of (1) an {\it entity-aware content selection component}, that leverages entity guidance to select salient sentences (\cref{ssec:content_selection}), and (2) an {\it abstract generation component} (\cref{ssec:abstract_generation}), that is trained with reinforcement learning to generate coherent and concise summaries (\cref{ssec:coherence}). Finally, we describe how the two components are connected to further improve the generated summaries (\cref{ssec:end2end}). 
\begin{figure}[t]
    \centering
    \includegraphics[width=\columnwidth,trim=0.55cm 0 0.1cm 0, clip]{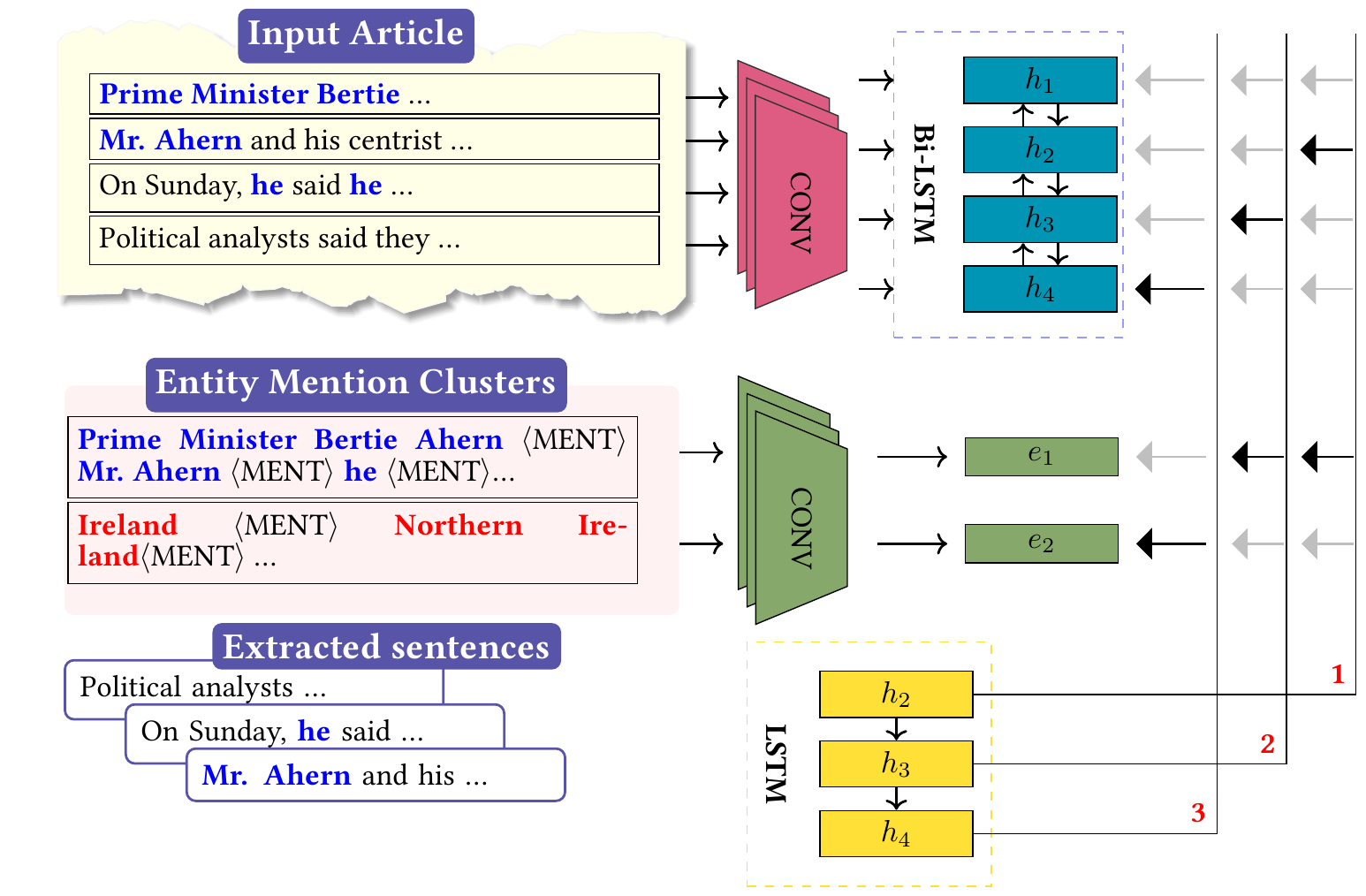}
    \vspace{-7mm}
    \captionof{figure}{ 
    Our proposed entity-aware content selector. Arrows denote attention, with darker color representing higher weights. 
    }
    \label{fig:extractor}
    \vspace{-4mm}
\end{figure}
\subsection{Entity-Aware Content Selection} 
\label{ssec:content_selection}
We design our content selection component to capture the interaction between entity mentions and the input article. 
Our model learns to identify salient content by aligning entity mentions and their contexts with human summaries. 
Concretely, we employ two encoders: one learns entity representations by encoding their mention clusters and the other learns sentence representations. 
A pointer-network-based decoder~\cite{vinyals2015pointer} selects a sequence of important sentences by jointly attending to the entities and the input, as depicted in~\cref{fig:extractor}.

\smallskip
\noindent \textbf{Entity Encoder.} 
We run off-the-shelf coreference resolution system from Stanford CoreNLP~\cite{manning-EtAl:2014:P14-5} on the input articles to extract entities, each represented as a cluster of mentions. Specifically, from each input article, we extract the coreferenced entities, and construct the mention clusters for all the mentions of each entity in that article. We also consider non coreferenced entity mentions as singleton entity mention clusters. Among all these mention clusters, for our experiments, we only consider salient entity mention clusters. We label clusters as ``salient" based on two rules: (1) mention clusters with entities appearing in the first three sentences of the article, and (2) top $k$ clusters containing most numbers of mentions. We experimented with different values of k and found that $k=6$ gives us the best set of salient mention clusters having an optimal overlap with entity mentions in the ground truth summary.



For each mention cluster, we concatenate mentions of the same entity as they occur in the input into one sequence, segmented with special tokens ($<$\textsc{ment}$>$). 
Finally, we get entity representations $\mathbf{e}_i$ for the $i$-th entity by encoding each cluster via a temporal convolutional model~\cite{kim-2014-convolutional}.



\smallskip
\noindent \textbf{Input Article Encoder.} 
For article encoding, we first learn sentence representations $\mathbf{r}_j$ by encoding words in the $j$-th sentence with another temporal convolutional model. Then, we utilize a bidirectional LSTM (biLSTM) to aggregate sentences into a sequence of hidden states $\bm{h}_j$. Both the encoders use a shared word embedding matrix to allow better alignment.



\smallskip
\noindent \textbf{Sentence Selection Decoder.} 
We employ a single-layer unidirectional LSTM with hidden states $\mathbf{s}_t$ to recurrently extract salient sentences. At each time step $t$, we first compute an entity context vector $\mathbf{c}^e_t$ based on attention mechanism~\cite{bahdanau2014neural}: 

\vspace{-5mm}
{\fontsize{9}{10}\selectfont
\begin{flalign}
\mathbf{c}^e_t = \sum_{i} a^e_{it} \mathbf{e}_i \\ 
\mathbf{a}^e_{t} = \text{softmax}(\mathbf{v}^e \tanh(\mathbf{W}^{e1} \mathbf{s}_t + \mathbf{W}^{e2} \mathbf{e}_i)) 
\end{flalign}}
%
where $\mathbf{a}_{t}^e$ are attention weights, $\mathbf{v}^{\ast}$ and $\mathbf{W}^{\ast \ast}$ denote trainable parameters throughout the paper. Bias terms are omitted for simplicity. 
We further use a \textit{glimpse operation} \cite{vinyals2015order} to compute a sentence context vector $\mathbf{c_t}$ as follows: 

\vspace{-3mm}
{\fontsize{9}{10}\selectfont
\begin{flalign}
\mathbf{c}_t = \sum_{j} a^h_{jt} \mathbf{W}^{h2} \mathbf{h}_j \\
\mathbf{a}^h_{t} = \text{softmax}(\mathbf{v}^h \tanh(\mathbf{W}^{h1} \mathbf{s}_t + \mathbf{W}^{h2} \mathbf{h}_j))
\end{flalign}}
where $\mathbf{a}_{t}^h$ are attention weights.
Finally, sentence extraction probabilities that consider both entity and input context are calculated as: 

\vspace{-3mm}
{\fontsize{9}{10}\selectfont
\begin{align}
\notag p(y_t^l | y_{1:t-1}^l) = \text{softmax}(\mathbf{v}^q \tanh(\mathbf{W}^{p1} \mathbf{s}_t + \mathbf{W}^{p2} \mathbf{c_t} \\ 
+ \mathbf{W}^{p3} \mathbf{c^e_t}))
\end{align}}
where the sentence $y_t^l$ with the highest probability is selected. The process stops when the model picks the end-of-selection token. 

\smallskip
\noindent \textbf{Selection Label Construction.} 
We train our content selection component with a cross-entropy loss: $-\sum_{(\mathbf{y}^l, \mathbf{x}) \in D} {\log{p(\mathbf{y}^l\, | \,\mathbf{x};\theta)}}$, here $\mathbf{y}^l$ are the ground truth sentence selection labels and $\mathbf{x}$ is the input article. $\theta$ denotes all model parameters. 

To acquire training labels for sentence selection, we collect positive sentences in the following way. 
First, we employ greedy search to select the best combination of sentences that maximizes ROUGE-2 F1~\cite{Lin:2003:AES:1073445.1073465} with reference to human summary, as described by~\newcite{zhou-etal-2018-neural-document}. 
We further include sentences whose ROUGE-L recall is above $0.5$ when each is compared with its best aligned summary sentence. 
In cases where no sentence is selected, we label the first two sentences from the article as positive. Our combined construction strategy selects an average of $2.96$ and $3.18$ sentences from New York Times and CNN/Daily Mail articles respectively.

\subsection{Abstract Generation with Reinforcement Learning}
\label{ssec:abstract_generation}

Our abstract generation component takes the selected sentences as input and produces the final summary. This abstract generator is a sequence-to-sequence network with attention over input~\cite{bahdanau2014neural}. The copying mechanism from \newcite{see-etal-2017-get} is adopted to allow out-of-vocabulary words to appear in the output. 

The abstract generator is first trained with maximum likelihood (ML) loss followed by additional training with policy-based reinforcement learning (RL). For ML training, we use teacher forcing algorithm~\cite{williams1995gradient}, to minimize the following loss: 



{\fontsize{9}{10}\selectfont
\begin{equation} \label{eq:lmloss}
\mathcal{L}_{\rm ml} = -\sum_{(\mathbf{y}, \mathbf{x}^{ext}) \in D} {\log{p(\mathbf{y}\, | \,\mathbf{x}^{ext};\theta)}}
\end{equation}
} 
where $D$ is the training set, $\mathbf{x}^{ext}$ are extracted sentences from our label construction. 

\medskip
\noindent \textbf{Self-Critical Learning.} 
Following \newcite{paulus2017deep}, we use the self-critical training algorithm based on policy gradients to use discrete metrics as RL rewards. 
At each training step, we generate two summaries: a {\it sampled} summary $\mathbf{y}^s$, obtained by sampling words from the probability distribution $p(\mathbf{y}^s | \,\mathbf{x}^{ext};\theta)$ at each decoding step, and a {\it self-critical baseline} summary $\hat{\mathbf{y}}$, yielded by greedily selecting words that maximize the output probability at each time step~\cite{rennie2017self}. 
We then calculate rewards based on the average of ROUGE-L F1 and ROUGE-2 F1 of the two summaries against that of the ground-truth summary, and define the following loss function:

\vspace{-4mm}
{\fontsize{9}{8}\selectfont
\begin{flalign} \label{eq:rlrloss}
\mathcal{L}_{\rm rl} = -\frac{1}{N^{\prime}} \sum_{(\mathbf{y}^s, \mathbf{x}^{ext}) \in D^\prime} (\mathbf{R}(\mathbf{y}^s) - \mathbf{R}(\hat{\mathbf{y}}))\log{p(\mathbf{y}^s | \,\mathbf{x}^{ext};\theta)}
\vspace{-8mm}
\end{flalign}
}
where $D^\prime$ represents set of sampled summaries paired with extracted input sentences and $N^\prime$ represents the total number of sampled summaries. {\small $\mathbf{R}(\mathbf{y})=\mathbf{R}_{\rm Rouge}({\mathbf{y}})= \frac{1}{2} \big({\mathbf{R}_{\rm Rouge-L}(\mathbf{y})+\mathbf{R}_{\rm Rouge-2}(\mathbf{y})}\big)$}, is the overall ROUGE reward for a summary $\mathbf{y}$.

\subsection{Rewards with Coherence and Linguistic Quality}
\label{ssec:coherence}

So far, we have described the \emph{two basic components} of our SENECA framework. 
As noted in prior work~\cite{liu2016not}, optimizing for an ngram-based metric like ROUGE does not guarantee improvement over readability of the generations. We thus augment our framework with additional rewards based on \emph{coherence} and \emph{linguistic quality} as described below.


\medskip
\noindent \textbf{Entity-Based Coherence Reward} ($\mathbf{R}_{\rm Coh}$). 
We use a separately trained {\bf coherence model} to score summaries and guide our abstract generator to produce more coherent outputs by adding a reward $\mathbf{R}_{\rm Coh}$ in the aforementioned RL training process. The new reward takes the following form:


\vspace{-2mm}
{\fontsize{9}{10}\selectfont
\begin{equation} \label{eq:cohreward}
\mathbf{R}(\mathbf{y}) =  \mathbf{R}_{\rm Rouge}({\mathbf{y}}) + \gamma_{\rm Coh} \mathbf{R}_{\rm Coh}({\mathbf{y}})
\end{equation}
\vspace{-3mm}
}

Here we show how to calculate $\mathbf{R}_{\rm Coh}$, to capture both entity distribution patterns and topical continuity. Since summaries are short, (e.g. $2.0$ sentences on average per summary in the New York Times data), 
we decide to build our coherence model on top of local coherence estimation for pairwise sentences. 
We adopt the architecture of neural coherence model developed by~\newcite{wu2018learning}, but train it with samples that enable coherence modeling based on entity presence and their context. Here we briefly describe the model, and refer the readers to the original paper for details. 

Given a pair of sentences $S_{A}$ and $S_{B}$, convolution layers first transform them into hidden representations, from which a multi-layer perceptron is utilized to yield a coherence score $\mathbf{Coh}(S_{A}, S_{B})\in [-1, 1]$. We train the model with hinge-loss by leveraging both coherent positive samples and incoherent negative samples: 

\vspace{-2mm}
{\fontsize{9}{10}\selectfont
\begin{flalign}
\notag \mathcal{L}(S_{A},S_{B}^+, S_{B}^-) = \max \{0, 1 + \mathbf{Coh}(S_{A}, S_{B}^+) \\
 - \mathbf{Coh}(S_{A}, S_{B}^-)\}  
\end{flalign}
} 
where $S_{A}$ is a target sentence, $S_{A}$ and $S_{B}^+$ is a positive pair, and $S_{A}$ and $S_{B}^-$ is a negative pair. 

Note that \newcite{wu2018learning} only consider {\it position} information for training data construction, i.e., $S_{A}$ and $S_{B}^+$ must be adjacent, and $S_{A}$ and $S_{B}^-$ are at most $9$ sentences away with $S_{B}^-$ randomly picked. We instead introduce two notable features to construct our training data.
In addition to being adjacent, we further constrain $S_{A}$ and $S_{B}^+$ to have at least one coreferred entity and that $S_{B}^-$ does not.
Since our initial experiments show that coherence model trained in this manner cannot discern pure repetition of sentences, e.g., simply duplicating words leads to higher coherence, we reuse the target sentences themselves as the negative pairs.

Finally, since this model outputs pairwise coherence scores, for a summary containing more than two sentences, we use the average of all adjacent sentence pairs' scores as the final summary coherence score. Summaries containing only one sentence get $0$ coherence score. We also conduct correlation study to show average aggregation works reasonably well (details in Supplementary).

\medskip
\noindent \textbf{Linguistic Quality Rewards} ($\mathbf{R}_{\rm Ref}$ \& $\mathbf{R}_{\rm App}$). 
We further consider two linguistically-informed rewards to further improve summary clarity and conciseness by penalizing (1) improper usage of referential pronouns, and (2) redundancy introduced by non-restrictive appositives and relative clauses. 

\medskip
\noindent \textit{Pronominal Referential Clarity.}
Referential pronouns occurring without the antecedents in a summary decreases its readability. For instance, a text with a pronoun ``\textit{they}" occurring before the required referred entity is introduced, would be less comprehensible. 
Therefore, at the RL step, we either penalize a summary with a reward of $-1$ for such improper usage, or give $0$ otherwise. In our implementation, we define improper usage as the presence of a third personal pronoun or a possessive pronoun before any noun phrase occurs. The new reward is written as $\mathbf{R}(\mathbf{y}) =  \mathbf{R}_{\rm Rouge}({\mathbf{y}}) + \gamma_{\rm Ref} \mathbf{R}_{\rm Ref}({\mathbf{y}})$.


\medskip
\noindent \textit{Apposition.} 
Next, we consider a reward to teach the model to use apposition and relative clause minimally, which improves summary conciseness. For this, we focus on the non-restrictive appositives and relative clauses, which often carry non-critical information~\cite{conroy2006back,wang-EtAl:2013:ACL20132} and can be automatically detected based on comma usage patterns. Specifically, a sentence contains a non-restrictive appositive if \begin{enumerate*}[label=\roman*)]
    \item it contains two commas, and
    \item the word after first comma is a possessive pronoun or a determinant~\cite{geva2019discofuse}.
\end{enumerate*}  
%
We penalize a summary with $-1$ for using non-restrictive appositives and relative clauses, henceforth referred to as apposition, or give $0$ otherwise. Similarly, we have the total reward as $\mathbf{R}(\mathbf{y}) =  \mathbf{R}_{\rm Rouge}({\mathbf{y}}) + \gamma_{\rm App} \mathbf{R}_{\rm App}({\mathbf{y}})$.

\subsection{Connecting Selection and Abstraction} 
\label{ssec:end2end}

Our entity-aware content selection component extracts salient sentences whereas our abstract generation component compresses and paraphrases them. Until this point, they are trained separately without any form of parameter sharing. We add an {\it additional step to connect these two networks} by training them together via the self-critical learning algorithm based on policy gradient (the same methodology as in~\cref{ssec:abstract_generation}). 

Following the Markov Decision Process formulation, at each time step $t$, our content selector generates a set of extracted sentences ($\hat{\mathbf{x}}^{ext}$) from an input article. Our abstract generator uses $\hat{\mathbf{x}}^{ext}$ to generate a summary. This summary, evaluated against the respective human summary, receives ROUGE-1 as reward (See~\cref{eq:rlrloss}). Note that the abstract generator, that has been previously trained with average of ROUGE-L and ROUGE-2 as reward to promote fluency, is not updated during this step. 
In this extra stage, if our content selector accurately selects salient sentences, the abstract generator is more likely to produce a high-quality summary, and such action will be encouraged. Whereas, action resulting in inferior selections will be discouraged.



\section{Experimental Setups}
\label{sec:experiments}

\noindent \textbf{Datasets and Preprocessing.}
We evaluated our models on two popular summarization datasets: New York Times (NYT)~\cite{sandhaus2008new} and CNN/Daily Mail (CNN/DM)~\cite{hermann2015teaching}. 
For NYT, we followed the preprocessing steps by \newcite{paulus2017deep} to obtain similar training ($588,909$), validation ($32,716$) and test ($32,703$) samples. Here, we additionally replaced author names with a special token. For CNN/DM, we followed the preprocessing steps in \newcite{see-etal-2017-get}, obtaining $287,188$ training, $13,367$ validation and $11,490$ testing samples.

For training our coherence model for CNN/DM, we used $890,419$ triples constructed from summaries and input articles sampled from the CNN/DM training set. Similarly for NYT, we sampled $884,494$ triples from NYT training set. For the validation and test set for the two models, we sampled roughly $10\%$ from the validation and test set of the respective datasets. Our coherence model for CNN/DM achieves 86\% accuracy and for NYT, 84\%. Additional evaluation for this model is reported in \cref{ssec:ecoh_eval}.



\medskip
\noindent \textbf{Training Details and Parameters.} 
We used a vocabulary of $50K$ most common words in the training set~\cite{see-etal-2017-get}, with $128$-dimensional word embeddings randomly initialized and updated during training. 
In the content selection component, for both entity and sentence encoders, we implemented one-layer convolutional network with $100$ dimensions and used a shared embedding matrix between the two. We employed LSTM models with $256$-dimensional hidden states for the input article encoder (per direction) and the content selection decoder~\cite{chen-bansal-2018-fast}. We used a similar setup for the abstract generator encoder and decoder. During ML training of both components, Adam~\cite{kingma2014adam} is applied with a learning rate $0.001$ and a gradient clipping $2.0$, and the batch size $32$. 
During RL stage, we reduced learning rate to $0.0001$~\cite{paulus2017deep} and set batch size to $50$. For our abstract generator, to reduce variance during RL training, we sampled 5 sequences for each data point and took an average over these samples using batch size 10. We set $\gamma_{\rm Coh}=0.01$, $\gamma_{\rm Ref}=0.005$, $\gamma_{\rm App}=0.005$ for NYT and CNN/DM with grid search on validation set. 

During inference, we adopted the trigram repetition avoidance strategy~\cite{paulus2017deep,chen-bansal-2018-fast}, with additional length normalization to encourage the generation of longer sequences as in \cite{gehrmann-etal-2018-bottom}.

\smallskip
\noindent \textbf{Baselines and Comparisons.} 
Besides baseline \textsc{Lead-3}, we compare with popular and existing state-of-the-art abstractive summarization models on NYT and CNN/DM datasets: 
(1) pointer-generator model with coverage~\cite{see-etal-2017-get} (\textsc{PointGen+cov}); 
(2) sentence rewriting model~\cite{chen-bansal-2018-fast} (\textsc{SentRewrite}); 
(3) RL-based abstractive summarization~\cite{paulus2017deep} (\textsc{DeepReinforce}); 
(4) bottom-up abstraction~\cite{gehrmann-etal-2018-bottom} (\textsc{BottomUp}); 
and (5) deep communicating agents-based summarization~\cite{celikyilmaz2018deep} (\textsc{DCA}).

We show comparison models' results reported as in the original publications. For significant tests, we run code released by the authors of \textsc{PointGen+cov} and \textsc{SentRewrite}, and by our implementation of \textsc{DeepReinforce} 
on both datasets for training and testing. Since we do not have access to model outputs from \newcite{paulus2017deep}, we re-implement their model, and achieve comparable ROUGE scores (e.g. on NYT, our ROUGE-1,2,L are 46.61, 29.76, and 43.46). For \textsc{BottomUp}, we obtain model outputs from the authors for both CNN/DM and NYT datasets, whereas for \textsc{DCA}, we acquire summaries only for CNN/DM dataset.

In addition to \textsc{SENECA} base model, which combines entity-aware content selection and RL-based abstract generation (with average of ROUGE-L F1 and ROUGE-2 F1 as reward), we also report results on its variants with additional rewards during abstract generator training. 
We further consider \textsc{SENECA} (i) without entities, and (ii) end-to-end trained but with sentence selection only, i.e., the abstract generator simply repeats the input.

\begin{figure}[t]
    \centering
    \includegraphics[scale=0.7]{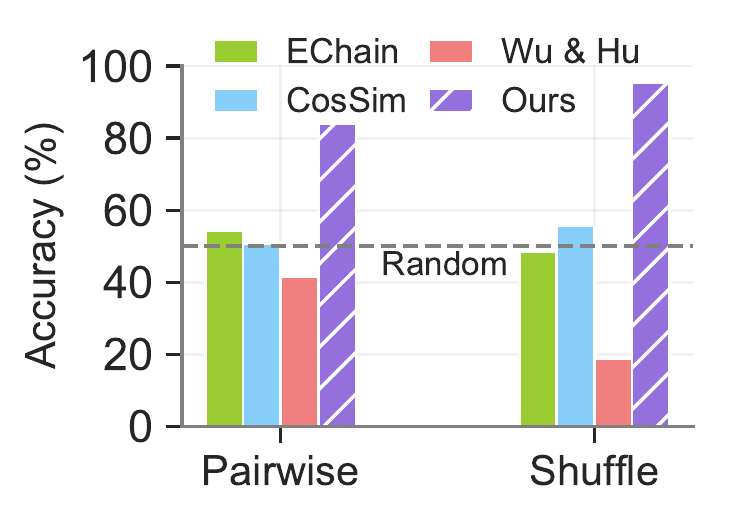}
    \vspace{-4mm}
    \captionof{figure}{ 
    Accuracy of our coherence model compared to different baselines and \newcite{wu2018learning} on \textsc{Pairwise} and \textsc{Shuffle} test sets. 
    }
    \label{fig:coherence_evaluation}
    \vspace{-3.5mm}
\end{figure}
\section{Results}
\label{sec:result}
In \cref{ssec:ecoh_eval}, we first evaluate our entity-based coherence model, which produces the coherence reward ($R_{\rm Coh}$). We then present automatic evaluation for summarization models on content, coherence, and linguistic quality (\cref{ssec:rouge_results}). 
We further discuss human evaluation results in \cref{ssec:human_eval}.

\subsection{Coherence Model Evaluation} 
\label{ssec:ecoh_eval}

We evaluate our entity-based coherence model on two tasks constructed from NYT test set: \textsc{Pairwise} and \textsc{Shuffle}. 
\textsc{Pairwise} is constructed as described in~\cref{ssec:abstract_generation} with equal number of positive pairs and negative pairs. 
\textsc{Shuffle} comprises of full summaries, where half are original summaries, and the other half contain their shuffled version. 

In~\cref{fig:coherence_evaluation}, we show a comparison of our model against a version trained based on the same amount of samples constructed as done by~\newcite{wu2018learning}. Also shown are two baselines: \textsc{EChain}, that labels a pair of sentences as more coherent if they have one or more entity mentions coreferred, and \textsc{CosSim}, that labels a pair of sentences with higher cosine similarity as more coherent. Our model yields significantly higher accuracy (greater than $80\%$) on both tasks than the comparisons, which is due to the difference in training data construction. \newcite{wu2018learning} only consider position information, whereas we capture entity-based coherence. The improvement in performance of the coherence model indicates the effectiveness of our training data construction in capturing multiple aspects of coherence.

\subsection{Automatic Summary Evaluation} 
\label{ssec:rouge_results}
\noindent \textbf{Results on NYT.} 
We first report the new state-of-the-art results for ROUGE-2 and ROUGE-L~\cite{Lin:2003:AES:1073445.1073465}, where our models outperform the previous best performing model \textsc{DCA}. Our \textsc{SENECA} models also outperform all comparisons on coherence score.  This indicates that our models' summaries not only contain more salient information but are also more coherent. 
\begin{table}[t]
\centering
\fontsize{9}{11}\selectfont
\setlength{\tabcolsep}{1.5mm}
\begin{tabular}{@{}lcccc@{}}
\toprule
\textbf{System} & \textbf{R-1} & \textbf{R-2} & \textbf{R-L} & \textbf{\textsc{Coh.}} \\ \midrule
\textsc{Human} & -	& -	& - & 0.79 \\ 
\textsc{Lead-3} & 32.59	& 16.49	& 29.17 & -\\ \hline
\textsc{PointGen+cov}$^{\dagger}$ &  41.06  & 25.71 &37.28  & 0.63  \\
\textsc{SentRewrite}$^{\dagger}$ &  42.19&	25.42&	38.74  & 0.32  \\ 
\textsc{DeepReinforce}$^\ast$~$^\dagger$ &  47.03 &	30.72 &	43.10  & 0.59 \\ 
\textsc{BottomUp}$^\ast$ & 47.38 &	31.23 &	41.81  & 0.56 \\ 
\textsc{DCA}$^\ast$ & {\bf 48.08} &	31.19	& 42.33  & - \\ \hline
{\bf Our Models}& 	& 	&  & \\
\textsc{SENECA} &  47.94 &	\textbf{\textit{31.77}} &	44.34 & 0.73 \\ 
\quad Input w/o Entity &  47.04 & 30.94 &	43.64  & 0.06 \\ 
\quad Sentence Selection Only & 39.97 & 22.49 &	35.68 &	0.30   \\
\quad + $R_{\rm Coh}$ & 47.57 & 31.28	& 44.03  & 0.75 \\ 
\quad + $R_{\rm Ref}$ &  47.57 & 31.22	& 43.92 & 0.70 \\ 
\quad + $R_{\rm App}$ &  \textit{48.05} & 31.71 & \textbf{\textit{44.60}} & 0.69  \\ 
\quad + $R_{\rm Coh}$+$R_{\rm Ref}$+$R_{\rm App}$ &  47.52 &	31.25 &	44.01 & \bf{0.76}  \\ 
\bottomrule
\end{tabular}
\vspace{-3mm}
\caption{
Results on NYT. Best results per metric are in bold. Best of our models are in italics. 
\textsc{SENECA} yields significantly higher R-1,2 and L than all comparisons except for \textsc{BottomUp} and \textsc{DCA}\footnotemark  (approximate randomization test ($p<0.0005$) ~\cite{edgington1969approximate}).
For \textsc{Coh.}, our best model is also significantly better (Welch's $t$-test, $p<0.05$).
$^\ast$: scores taken from original paper. $^{\dagger}$: significant test done on outputs by running code released by their authors or our implementation. 
}
\label{tab:nyt-rouge}
\vspace{-3mm}
\end{table}
\footnotetext{Significant test was not performed with \textsc{BottomUp} as they use different test split than \newcite{paulus2017deep}, and with \textsc{DCA}, since their outputs are unavailable on NYT.}

Amongst our models, the base \textsc{SENECA} model reports higher ROUGE and coherence score ($0.73$) compared to the version without entity as input. This demonstrates the importance of entity guidance during content selection as well as abstract generation. \textsc{SENECA} model trained with Apposition reward ($R_{\rm App}$) reports the highest ROUGE-L ($44.60$), but shows a drop in the coherence score to $0.69$. Whereas, \textsc{SENECA} with all rewards $R_{\rm Coh}$+$R_{\rm Ref}$+$R_{\rm App}$ reports the highest coherence score of $0.76$ and a drop in ROUGE-L ($44.01$). 
 \\

\begin{table}[t]
\centering
\fontsize{9}{11}\selectfont
\setlength{\tabcolsep}{1.5mm}
\begin{tabular}{@{}lcccc@{}}
\toprule
\textbf{System} & \textbf{R-1} & \textbf{R-2} & \textbf{R-L} & \textbf{\textsc{Coh.}} \\\midrule
\textsc{Human} & -	& -	& - & 0.55  \\
\textsc{Lead-3} & 40.23	& 17.52	& 36.34 & -  \\ \hline
\textsc{PointGen+cov}$^\ast$ $^{\dagger}$ & 39.53 &	17.28 &	36.38  & 0.19  \\ 
\textsc{SentRewrite}$^\ast$ $^{\dagger}$ &  40.88 &	17.80	& 38.54  & 0.41  \\ 
\textsc{DeepReinforce}$^\ast$ $^{\dagger}$ &  41.16 & 15.75 & {\bf 39.08}  & -0.21 \\
\textsc{Bottom-Up}$^\ast$ $^{\dagger}$ & 41.22&	18.68 &	38.34  & 0.47 \\
\textsc{DCA} $^{\dagger}$ & 40.72 & \textbf{19.02} & 37.85 & 0.30 \\\hline
{\bf Our Models}& 	& 	&  & \\
\textsc{SENECA} &  \textbf{\textit{41.52}} &	18.36 &	38.09 & -0.12 \\ 
\quad Input w/o Entity &  38.18	& 16.98 & 34.20 & -0.15  \\ 
\quad Sentence Selection Only & 41.43	& \textit{18.84}	& 37.73 & -0.06 \\ 
\quad + $R_{\rm Coh}$ & 41.04	& 16.98 & 38.08 & 0.52   \\ 
\quad + $R_{\rm Ref}$ & 41.35	& 16.98 &	\textit{38.25} & -0.30   \\ 
\quad + $R_{\rm App}$ &  41.38 & 17.22 &	\textit{38.43} & 0.00   \\ 
\quad + $R_{\rm Coh}$+$R_{\rm Ref}$+$R_{\rm App}$ &  40.71 &	16.68 &	38.07 & \bf{0.63}   \\
\bottomrule
\end{tabular}
\vspace{-3mm}
\caption{Results on CNN/Daily Mail. 
Our best results are achieved by the variant with sentence selection only, yielding significantly better R-1 and R-2 than all comparisons except \textsc{Bottom-Up} and \textsc{DCA} (Welch's $t$-test, $p<0.05$).
For \textsc{Coh.}, our best model is also significantly better (Welch's $t$-test, $p<0.05$).
We re-evaluate \textsc{DCA} output against full length human summaries. 
$^\ast$: scores taken from original paper. $^{\dagger}$: significant test done on outputs by running code released by their authors or our implementation. 
}
\label{tab:cnn-rouge}
\vspace{-4mm}
\end{table}

\smallskip
\noindent \textbf{Results on CNN/DM.} 
Since CNN/DM summaries are more extractive than that of NYT~\cite{grusky-etal-2018-newsroom}, all \textsc{SENECA} models produce higher ROUGE-1 scores with \textsc{SENECA} base model achieving the highest, outperforming the previous best performing models on ROUGE-1. We also report the highest coherence score ($0.63$), which is even higher than that reported on human summaries ($0.55$). Since CNN/DM gold summaries are comprised of concatenated human written highlights for input articles, they are about the same topic and are cohesive, but lack entity-based coherent structure, for instance fewer entities get coreferred in subsequent sentences. Therefore, our coherence evaluation, which tests for entity-based coherence, gives lower coherence score to CNN/DM gold summaries.

Additionally for CNN/DM, the generated summaries sometimes contain less relevant words, e.g. stopwords, at the end of the summaries. This effect however improves the ROUGE scores while adversely affecting the coherence scores of the summaries. Training with additional coherence reward alleviates the issue by appending additional sentences, thereby improving overall coherence.     

Amongst our models, we observe the overall trend to be similar to that in NYT results. Our base \textsc{SENECA} model reports higher ROUGE and coherence score compared to \textsc{SENECA} without entity input, again, indicating the usefulness of entity information. \textsc{SENECA} model with all rewards also yields the highest coherence score of $0.63$, whereas \textsc{SENECA} model with coherence reward performs considerably better on ROUGE-L with a drop in coherence score. 


\begin{table}[t]
\hspace{-1mm}
\fontsize{9}{10}\selectfont
\setlength{\tabcolsep}{0.3mm}
\begin{tabular}{@{}lccccccc@{}}
\toprule
System & \multicolumn{3}{c}{NYT} & & \multicolumn{3}{c}{CNN/DM} \\
 & \textbf{Ref.} & \textbf{RelCl.}  & \textbf{App.} & & \textbf{Ref.}  & \textbf{RelCl.}  & \textbf{App.} \\ \midrule
\textsc{Human} & 0.11 & 6.67 & 4.83 & &0.13 & 2.80 & 1.12 \\ \hdashline
\textsc{PointGen + cov} & 0.15 & 1.92 & 1.26 & & 0.18 & 0.63 & 0.75 \\
\textsc{SentRewrite} & 0.13 & 0.75 & 0.68 & & 0.11 & 0.28 & 0.19\\
\textsc{DeepReinforce} & 0.18 & 0.40 & 0.50 & & 0.19 & 0.45 & 0.10\\
\textsc{BottomUp} & 0.12  & 0.62 & 0.54 & & 0.10 & 0.23 & 0.09 \\
\textsc{DCA} & - & - & - & & 0.12 & 0.26 & 0.13 \\
\textsc{SENECA} & 0.13 & 1.15 & 0.68 & &0.10 & 0.22 & 0.06\\
\quad w/o entity & 0.21 & 1.24 & 0.70 & &0.11 & 0.33 & 0.10\\
\quad + $R_{\rm Coh}$ & 0.12 & 1.15 & 0.71 & & \textbf{0.06} & 0.07 & 0.05\\
\quad + $R_{\rm Ref}$ & \textbf{0.10} & 1.20 & 0.72 & &0.07 & 0.11 & 0.04\\
\quad + $R_{\rm App}$ & 0.13 & 0.65 & \textbf{0.42} & &0.09  & \textbf{0.01} & \textbf{0.01}\\
\quad + $R_{\rm Coh}$+$R_{\rm Ref}$+$R_{\rm App}$  & 0.12 & 0.94 & 0.59 & &0.07 & 0.04 & \textbf{0.01}\\ \bottomrule
\end{tabular}
\vspace{-2mm}
\caption{$\%$ of system summaries improperly using referential pronouns (Ref.), or containing relative clauses (RelCl.) or appositives (App.) (lowest values in bold). Lower values indicate better performance.
}
\label{tab:nyt-reclarity}
\vspace{-4mm}
\end{table}
\smallskip
\noindent \textbf{Linguistic Quality Evaluation.}
We further do a preliminary evaluation of system summaries on two important linguistic qualities: clarity and conciseness. 
To measure {\it clarity}, we focus on the percentage of summaries that improperly use referential pronouns (Ref.), defined as third person pronouns or possessive pronouns being used before any noun phrase. Similarly, to measure {\it conciseness}, we report how often summaries contain at least one non-restrictive relative clause (RelCl.) or non-restrictive appositives (App.). For model summaries, we report these measures in reference to the respective gold summaries. Lower values are better.

As can be seen from Table~\ref{tab:nyt-reclarity}, our models report the least percentage of improper usage of referential pronouns. Particularly on NYT, the model trained with $R_{\rm Ref}$ reward makes much fewer mistakes in this category. Similarly, adding $R_{\rm App}$ reports the least amount of usage of relative clause or apposition, making summaries easier to read.

\subsection{Human Evaluation} 
\label{ssec:human_eval}
Human evaluation is conducted to analyze the informativeness and readability of the summaries generated by our models. 
We randomly select $30$ articles from the NYT test set and ask three proficient English speakers to rate summaries generated by \textsc{PointGen+cov}, \textsc{DeepReinforce}~\cite{paulus2017deep}, 
our \textsc{SENECA}, and \textsc{SENECA} + $R_{\rm Coh}$,\footnote{To clearly discern the improvement in coherence in the summaries after adding coherence reward without introducing effects of additional rewards, we did not consider \textsc{SENECA} model with all rewards for human evaluation.} along with \textsc{Human} written summaries. 
Each rater reads the article and scores the summaries against each other on a Likert scale of $1$ (worst) to $5$ (best) on the following three aspects: 
\textbf{informativeness}---whether summary covers salient points from the input, 
\textbf{grammaticality}, 
and \textbf{coherence}---whether summary presents content and entity mentions in coherent order. 
Detailed guidelines with sample ratings and explanation are shown in the Supplementary. 
\begin{table}[t]
\centering
\fontsize{9}{11}\selectfont
 \setlength{\tabcolsep}{0.9mm}
  \centering
    \begin{tabular}{lccc}
        \toprule
        & {\bf Inf.} & {\bf Gram.} & {\bf Coh.} \\
        \midrule
        \textsc{Human} & 4.58 & 4.42 & 4.16    \\ \hdashline
        \textsc{PointGen+cov} & 3.63 & \textbf{4.47}  & 3.89    \\  
        \textsc{DeepReinforce} & 3.63 & 3.21 & 3.16    \\  
        \textsc{SENECA} & 4.26 & 4.11 & 4.05  \\
        \textsc{SENECA} + $R_{\rm Coh}$ & \textbf{4.32$^{\ast}$} & 4.10 & \textbf{4.11$^{\ast}$}   \\
        \bottomrule
    \end{tabular}
    \vspace{-2mm}
    \caption{
    Human evaluation on informativeness (Inf.), grammaticality (Gram.) and coherence (Coh.). Best results among systems are in bold, with significance reported over other comparisons with $\ast$ (approximate randomization test ~\cite{edgington1969approximate}, $p<0.0005$). The Krippendorf's $\alpha$ for the three aspects are $0.46$, $0.64$, and $0.48$.}
  \label{tab:human-eval}
  \vspace{-3mm}
\end{table}

Table~\ref{tab:human-eval} shows that our model \textsc{SENECA} + $R_{\rm Coh}$ ranks significantly higher on informativeness as well as coherence, reaffirming our observations from automatic evaluation. Surprisingly, \textsc{SENECA} + $R_{\rm Coh}$ ranks higher on informativeness even when compared to \textsc{SENECA}, which reports higher on ROUGE (see Table~\ref{tab:nyt-rouge}). 
Through manual inspection, we find that \textsc{SENECA} model often generates summaries whose second or third sentence misses the subject,
whereas \textsc{SENECA} + $R_{\rm Coh}$ tends to avoid this problem. 
%
%
Though not significant, \textsc{PointGen+cov} ranks higher on grammaticality than \textsc{SENECA} + $R_{\rm Coh}$. We believe this is due to the fact that \textsc{SENECA} + $R_{\rm Coh}$ learns to merge sentences from the input article while making some grammatical errors. We further show sample summaries in Figure~\ref{fig:sample-outputs}. 




\section{Related Work}
\label{sec:relatedwork}

\noindent \textbf{Neural Abstractive Summarization.} 
Two-step abstractive summarization approaches have become popular in recent years, where the two steps, content selection and abstraction, are conveniently separated from each other. In these approaches, salient content is first identified, usually at sentence-level~\cite{hsu-etal-2018-unified,li-etal-2018-guiding,tan-etal-2017-abstractive,chen-bansal-2018-fast} or phrase-level~\cite{gehrmann-etal-2018-bottom}, followed by abstract generation. However, prior work mainly focuses on improving the informativeness of abstractive summaries, e.g.\ copying and coverage mechanisms~\cite{see-etal-2017-get}, and reinforcement learning methods optimizing on ROUGE scores~\cite{paulus2017deep}. Coherence and other aspects of linguistic quality that capture the overall readability of summaries are largely ignored. In this work, besides informativeness, we also aim to improve upon these aspects of summaries.

\begin{figure}[t]
\centering
\fontsize{9}{10}\selectfont
\setlength{\tabcolsep}{0.8mm}{
	\begin{tabular}{p{75mm}}
    \hline
    \vspace{0.1mm}
    \textbf{Human}: \\
    New Jersey Legislature recommends 0 ways to overhaul system that has produced highest property taxes in nation; plan includes 0 percent reduction in property taxes to most homeowners through direct tax credits; {\color{blue!70} \textbf{will place annual limit on property tax increases; will revise financing of education and end special financing of state's poor districts}}
    \\
    \hline
    \vspace{0.1mm}
     \textbf{PointGen+cov}:\\
     new jersey legislature, after more than three decades of complaints about soaring property taxes and three months of hearings about ways to reduce them, designed to overhaul system that has produced highest property taxes in nation. recommendations included 0 percent reduction in property taxes to most of state's homeowners in form of direct tax credits
     \\
	
	\hline
	\vspace{0.1mm}
	\textbf{DeepReinforce}: \\
	new jersey legislature, after more than three decades of complaints about soaring property taxes and three months of hearings about ways to reduce them. unveils 0 proposals designed to overhaul system that has highest property taxes in nation. recommendations include 0 percent reduction in property taxes to most of state's and of direct tax credits
	\\
	
	\hline
	\vspace{0.1mm}
	\textbf{\textsc{SENECA}}:\\
	new jersey legislature unveils 0 proposals designed to overhaul system that has produced highest property taxes in nation. recommendations included 0 percent reduction in property taxes to most of state's homeowners in form of direct tax credits
	\\
	
    \hline
    \vspace{0.1mm}
    \textbf{\textsc{SENECA} + $\mathbf{R}_{\rm Coh}$}:\\
    new jersey legislature, unveiled 0 proposals designed to overhaul system that has highest property taxes in nation. recommendations included 0 percent reduction in property taxes to most of state's homeowners in form of direct tax credits. {\color{green!75!blue} \textbf{other parts of plan would place limit on annual property tax increases and revise way state pays for public education, ending special financing given to state's poor districts}}
    \\
    \hline
    
	\end{tabular}
	}
	\vspace{-1mm}
	\caption{
	Sample summaries for an NYT article. 
	Our model with coherence reward overlaps the most with human summary ({\color{green!75!blue}\textbf{green}} is ours, {\color{blue!70}\textbf{blue}} denotes human). }
	
\label{fig:sample-outputs}
\vspace{-2mm}
\end{figure}

\smallskip
\noindent \textbf{Role of Entities and Coherence in Summarization.} 
Entities in a text carry useful contextual information ~\cite{nenkova2008entity} and therefore play an important role in multi-document summarization~\cite{li-etal-2006-extractive} and event summarization for selecting salient sentences~\cite{li2015reader}. Moreover, entity mentions connecting sentences have also been used to extract non-adjacent yet coherent sentences~\cite{siddharthan2011information,parveen-etal-2016-generating}. 
For abstractive summarization, \newcite{amplayo2018entity} find it beneficial to leverage entities that are linked to existing knowledge bases. Unfortunately, it fails to capture the entities that do not exist in these knowledge bases.

Grammar role-based entity transitions have been widely employed to model coherence in text generation tasks~\cite{barzilay-lee-2004-catching,lapata2005automatic,barzilay2008modeling,guinaudeau-strube-2013-graph,tien-nguyen-joty-2017-neural}, which often requires intensive feature engineering. Neural coherence models~\cite{mesgar-strube-2018-neural,li-hovy-2014-model} have, therefore, gained popularity due to their end-to-end nature. 
However, coherence has mainly been investigated in extractive summarization systems~\cite{alonso2003integrating,christensen2013towards,parveen-etal-2015-topical,wu2018learning}. To the best of our knowledge, we are the first to leverage entity information to improve coherence for neural abstractive summarization along with other important linguistic qualities.

\section{Conclusion}
\label{sec:conclusion}
We present an entity-driven summarization framework to generate informative and coherent abstractive summaries. An entity-aware content selector chooses salient sentences from the input article and an abstract generator produces a coherent abstract. Linguistically-informed guidance further enhances conciseness and clarity, thus improving the summary quality.
Our model obtains the new state-of-the-art ROUGE scores on the NYT and CNN/DM datasets. Human evaluation further indicates that our system produces more coherent summaries than other popular methods.

\section*{Acknowledgements}
This research is supported in part by National Science Foundation through Grants IIS-1566382 and IIS-1813341, and by the Office of the Director of National Intelligence (ODNI), Intelligence Advanced Research Projects Activity (IARPA), via contract \# FA8650-17-C-9116. The views and conclusions contained herein are those of the authors and should not be interpreted as necessarily representing the official policies, either expressed or implied, of ODNI, IARPA, or the U.S. Government. The U.S. Government is authorized to reproduce and distribute reprints for governmental purposes notwithstanding any copyright annotation therein. We thank the anonymous reviewers for their constructive suggestions. We also thank Sebastian Sehrmann for sharing their outputs on NYT and CNN/DM and Asli Celikyilmaz for sharing their summaries on CNN/DM.

\bibliography{references}
\bibliographystyle{acl_natbib}
\appendix
\section{Supplemental Material}
\label{sec:supplemental}
\subsection{Entity-based Coherence Model}
In ~\cref{ssec:coherence}, we explain that, since the coherence model outputs pairwise coherence scores, for a summary containing more than two sentences, we use the average of all adjacent sentence pairs' scores as the final summary coherence score. Here, we explain the correlation study we conducted to show that the average aggregation performed in this case, works reasonably well.

We first sample 300 such NYT summaries from our test set. Considering the original summary containing two sentences as a positive sample, for each summary, we construct a negative sample by reversing its original sentence order. If a summary contains more than two sentences, we further construct a ``middle'' sample by shuffling its sentence order such that it is different from positive and negative samples. We give each positive sample a coherence score of 3 (perfect score), a middle sample 2, and a negative sample 1 (worst score). We call this ideal coherence score. For each of these samples, we also get the coherence score given by our model using average aggregation to score each sample summary. Finally, we calculate the Spearman rank-order correlation coefficient between the ideal coherence score and our model score using SciPy\footnote{https://docs.scipy.org/doc/scipy/reference/index.html}. The final correlation coefficient is 0.87, with p-value 1.43e-237, which shows that the two coherence scores are highly correlated. This implies that aggregated coherence score for a summary can be used as a reliable proxy for its overall coherence score.   

\subsubsection{Entity-based Coherence Model Evaluation} 
To investigate the robustness of our coherence model, we constructed several test sets to evaluate its performance on different dimensions of coherence, which we describe in detail here. We also considered three baselines as a control for some of these dimensions. Each test instance in our test sets had two sentences, except for the \textsc{Shuf.}. We describe our test data construction and baselines as: \\
\textsc{Pair}: We construct this test set in the same way as the training data for our model (as in ~\cref{ssec:ecoh_eval}).\\
\textsc{Conn.}: To test whether model learns to discriminate between proper and improper usage of discourse connectors, we replace the discourse connectors connecting two sentences with a connector of different type in the incoherent pairs (We borrow the list of discourse connectors and rules from \newcite{geva2019discofuse}). We implement \textsc{Top20Con}, as a baseline for this test set, which labels a pair of sentences using top 20 most frequently used discourse connectors as more coherent.\\
\textsc{Ref. and Ent.Rep.}: Here, we want to detect whether model learns to discriminate between proper and improper usage of pronouns and entity repetition. For pronouns (Ref.), we swap the pronominal entity mention  with the respective nominal entity mention to construct incoherent pairs. For entity repetition (Ent.Rep.), we replace the pronominal entity mention in the second sentence with the respective nominal entity mentioned in first sentence to construct incoherent pairs. For both of these respective test sets, we use \textsc{EChain} as a baseline, that labels a pair of sentences as more coherent if they have one or more entity mentions co-referred.\\
\textsc{Over.} To confirm that model learns that coherent pair of sentences share topical content, we force incoherent sentence pairs to have no content overlap. For this, we consider \textsc{CosSim}, that labels a pair of sentences with higher cosine similarity as more coherent.\\
\textsc{Shuf.} To get the final coherence score for a summary, we take an average over the coherence scores of all the adjacent sentence pairs in a summary. Therefore, to reliably test model's performance on test instances containing more than two sentence, considering a human written summary as coherent, we shuffle sentences of each summary to construct respective incoherent pairs.

\begin{table}[h]
\centering
\small
\fontsize{9}{10}\selectfont
\setlength{\tabcolsep}{0.6mm}
\begin{tabular}{lcccccc}
\toprule
\textbf{Model} & \textsc{Pair.} & \textsc{Conn.}  & \textsc{Ref.} & \textsc{Ent.Rep.}  & \textsc{Over.}  & \textsc{Shuf.}  \\ \midrule
\textsc{Random} & 50.0 &  50.0 & 50.0 & 50.0 & 50.0 & 50.0 \\ \hdashline
\textsc{Top20Con} &  48.6  &  49.5 & - & - & - & 51.6 \\
\textsc{EChain} & 54.5 &  - & 61.9 & 43.1 & - & 48.5 \\
\textsc{CosSim} & 50.7 &  - & - & - & 79.2 & 55.7 \\
Wu \& Hu & 41.5  & 50.6  & 42.1 & 34.6 & 52.5 & 19.0  \\
\bf \textsc{Ours}  & \textbf{84.1}  & \textbf{73.0} & \textbf{94.2} & \textbf{93.4} & \textbf{85.7} & \textbf{95.4} \\ \bottomrule
\end{tabular}
\caption{Accuracy of Entity-based Coherence model compared to different baselines and \newcite{wu2018learning}. The listed test sets are constructed from NYT test set and contain 1000 samples each.}
\label{tab:coherence_evaluation}
\end{table}

Table ~\ref{tab:coherence_evaluation} lists the performance of our coherence model and ~\newcite{wu2018learning} on aforementioned test sets, along with four baseline systems. Our model achieves $95.4\%$ accuracy on \textsc{Shuf.}, which implies that it accurately captures overall coherence of a summary. Moreover, we find that our model performs significantly better on all other test sets too, compared to often worse than random performance of \newcite{wu2018learning} model on the same. We attribute this to the fact that \newcite{wu2018learning} data construction can not capture different dimensions of coherence. We also find our model to perform better than the three baselines on respective datasets. For instance, on \textsc{Conn.}, our model performs significantly better than \textsc{Random} and \textsc{Top20Con} indicating that it is not overfitting to the data, instead learns the pattern of discourse connector usage. 

\subsection{Human Evaluation Guideline}
In our human evaluation, each human annotator is presented with 30 news articles. The annotators are asked to evaluate five summaries for each article on the following aspects on scale of 1 - 5 (1 being very poor and 5 being very good). More detailed instructions are in Table \ref{tab:human_eval}.
\begin{itemize}
    \item \textbf{Informativeness}: whether the summary provides enough and necessary content coverage from the input article.
    \item \textbf{Grammaticality}: whether the summary has no obvious grammatically incorrect sentences (e.g., fragments, missing components) that make the text difficult to read.
    \item \textbf{Coherence}: weather the summary is coherent and well-organized.
\end{itemize}

\newpage
\begin{table*}
	\fontsize{10}{11}\selectfont
    \centering
    \begin{tabular}{lp{120mm}}
         \toprule
         \multicolumn{2}{c}{\textbf{Informativeness:}} \\
         \midrule
         \rowcolor{lightgray!30}
          1 & Not relevant to the article \\
          & e.g., \textit{"less than three months after state voters defeat proposal that would have outlawed most abortions . legislators , introduced scaled-back version"} \\
          
          \rowcolor{lightgray!30}
          3 & Relevant, but misses the main point of the article \\
          & e.g., \textit{"vermont supreme court rules isabella miller-jenkins has two mothers . The Vermont ruling , he added , illustrates that same-sex marriage or civil unions will inevitably clash with other states ."} \\
          \rowcolor{lightgray!30}
          5 &  Successfully captures the main point of the article \\
          & e.g., \textit{"vermont supreme court rejected host of arguments from isabella's biological mother lisa miller that her former lesbian partner janet jenkins should be denied parental rights . decision conflicts with one from court in virginia , where miller and her daughter , who is 0 , now live"} \\
          \midrule
         \multicolumn{2}{c}{\textbf{Grammaticality:}} \\
         \midrule
         \rowcolor{lightgray!30}
         1 & Summary contains fragments, missing components, extra punctuations \\
         & e.g., \textit{"court rejects host of arguments from isabella 's biological mother , lisa miller , that her former lesbian partner janet jenkins should be denied parental rights . miller and to , to"} \\
         \rowcolor{lightgray!30}
         3 & Relatively minor grammatical errors \\
         & e.g., \textit{"isabella miller-jenkins , vermont supreme court , rules rejects host of arguments from isabella 's biological mother , lisa miller ."} \\
         \rowcolor{lightgray!30}
         5 & Correct Grammar. \\
         & e.g., \textit{"vermont supreme court rejected host of arguments from isabella's biological mother lisa miller that her former lesbian partner janet jenkins should be denied parental rights ."} \\

          \midrule
         \multicolumn{2}{c}{\textbf{Coherence:}} \\
         \midrule
         \rowcolor{lightgray!30}
         1 & Completely incoherent with no across sentence organization \\
         & e.g., \textit{"isabella miller-jenkins , vermont supreme court , isabella miller-jenkins , vermont supreme court."} \\
          \rowcolor{lightgray!30}
         3 & Understandable, but sentences can be slightly re-ordered for it to read better \\
         & e.g., \textit{"vermont supreme court rejects arguments from lisa miller , biological mother of 0-year-old isabella miller-jenkins , that her former lesbian partner , janet jenkins , should be denied parental rights . isabella was born in virginia in 0 , after miller was impregnated with sperm from anonymous donor whom jenkins helped select. decision conflicts with one from court in virginia , where miller and her daughter now live ."} \\
         \rowcolor{lightgray!30}
         5 & Completely coherent. \\
         & e.g., \textit{"vermont supreme court rejected host of arguments from isabella 's biological mother lisa miller that her former lesbian partner janet jenkins should be denied parental rights . decision conflicts with one from court in virginia , where miller and her daughter , who is 0 , now live"} \\
         \bottomrule
    \end{tabular}
    \caption{Sample summaries with explanations on human evaluation aspect scales.}
    \label{tab:human_eval}
\end{table*}

\subsection{Sample Output}

We include 4 complete sample outputs for different models in Figures \ref{fig:nyt_sample1} to \ref{fig:cnndm_sample2}. 

\begin{figure*}[t]
    \fontsize{10}{11}
    \centering
    \setlength{\tabcolsep}{0.8mm}

	\begin{tabular}{|p{160mm}|}
    \hline
    \large{\textbf{\textsc{Article}}}: \\
    \vspace{-3mm}
    A Backyard Mayan Temple May Be Doomed. WHEN Barbara Winston had a stone replica of a Mayan temple built in her backyard, it was meant to give her a sense of peace. Instead, it started a war. In August 0, Mrs. Winston, who lives on a 0-acre estate with her husband, Bruce, had a stonecutter erect a \$ 0,0 replica of Guatemala's Temple of the Great Jaguar at the rear of their property. While the temple's scale is one-seventeenth of the original, it is by no means small, with granite blocks forming three staircases rising nine feet to a rectangular platform. It is also 0 feet from the horse barn and riding ring of the Winstons' neighbor, Diane Lewis , who complained to the Town of Bedford that the temple detracted from the enjoyment of her property. In September, the Zoning Board of Appeals rejected the Winstons' request for a variance and said that the temple was a structure requiring permits and a 0-foot setback from adjacent property. It questioned why the Winstons, with 0 acres to choose from , put the temple so close to Ms. Lewis's property. The ruling left the Winstons with three choices: tear down the temple, move it , or fight. 
    (...)
    In their lawsuit, the Winstons argue that the temple was not a structure as defined in the ordinance. Even if it were, they said, the town should grant the variance because of the temple's spiritual significance. Joel Sachs, the town attorney, said the temple clearly met the definition of a structure . (...) \\
    \hline
    \textbf{\textsc{Human}}: \\
    Bruce and Barbara Winston file suit against Zoning Board in Bedford, NY, after board \colorbox{green!45}{rejects their request for variance} for stone replica of Mayan temple that couple had built on their 0-acre estate. board, after complaint from neighbor Diane Lewis, says temple is structure that requires permits and 0-foot setback from adjacent property
    \\
    \vspace{-1mm}
     \textbf{\textsc{PointGen + cov}}:\\
    arbara winston, who lives on 0-acre estate with her husband, bruce, has stonecutter erect \$ 0,0 replica of guatemala's temple of great jaguar at rear of their property. it is by no means small, with granite blocks forming three staircases rising nine feet to rectangular platform
     \\
     \vspace{-1mm}
	\textbf{\textsc{DeepReinforce}}: \\
    article on barbara winston, \$ 0,0 replica of guatemala's temple of great jaguar at rear of their property. of zoning board of appeals, to request for variance and of that temple, to structure requiring permits and 0-foot from adjacent property.
	\\
	\vspace{-1mm}
    \textbf{\textsc{SENECA}}: \\
    winston, who lives on 0-acre estate with her husband, bruce. erect \$ 0,0 replica of guatemala's temple of great jaguar at rear of their property. zoning board of appeals \colorbox{green!45}{rejects winstons' request for variance} and says that temple, is structure requiring permits and 0-foot setback from adjacent property
	\\
	\vspace{-1mm}
    \textbf{\textsc{SENECA} + $R_{\rm Coh}$}:\\
    winston, who lives on 0-acre estate with her husband, bruce, has stonecutter erect \$ 0,0 replica of guatemala's temple of great jaguar at rear of their property. zoning board of appeals \colorbox{green!45}{rejects winstons' request for variance} and says that temple is structure requiring permits and 0-foot setback from adjacent property

    \\
    \hline
	\end{tabular}
	\vspace{2mm}
	\caption{
	Sample summaries for an NYT article. Our models capture \colorbox{green!45}{salient information} which is missed by comparisons. Numbers are replaced with ``$0$".
   }
\label{fig:nyt_sample1}
\end{figure*}


\begin{figure*}[t]
    \fontsize{10}{11}
    \centering
    \setlength{\tabcolsep}{0.8mm}

	\begin{tabular}{|p{160mm}|}
    \hline
    \large{\textbf{\textsc{Article}}}: \\
    \vspace{-3mm}
    TV Ad for Pfizer's Painkiller Criticized by Consumer Group. A new television commercial for Pfizer's painkiller Celebrex that has attracted attention for both its length and innovative marketing approach is now also being criticized for its message. Public Citizen, a consumer group, asked the Food and Drug Administration yesterday to ban the commercial, charging that it gives consumers a false impression that the prescription drug has no more safety risk than some other painkillers. Celebrex is in the same class of drugs as the Merck pill Vioxx, which was withdrawn in 0 because of its link to cardiovascular problems. At that time, the F.D.A. also asked Pfizer to suspend its television advertising for Celebrex, and the company complied. The new Celebrex ad, which can also be seen on a Pfizer Web site celebrex.com, represents a return to television for the product after a hiatus of more than two years. It was broadcast for the first time last Monday on World News With Charles Gibson on ABC. It was two and a half minutes rather than the usual 0 seconds , and was the only ad during last Monday's program. Mr. Gibson announced that the new format with less advertising would be repeated on several Mondays this month. Pfizer has also bought last night's and next Monday's World News broadcast, but had said all along that it would not disclose the content of those ads until they were broadcast. Last night's advertisement involved smoking cessation and directed viewers to a Web site, www.mytimetoquit.com that links to information about Chantix, Pfizer's prescription stop-smoking drug. (...) \\
    \hline
    \textbf{\textsc{Human}}: \\
    New television commercial for Pfizer's painkiller Celebrex that has attracted attention for its length and innovative marketing approach is being criticized for its message; Public Citizen asks Food and Drug Administration to ban commerical, charging that it gives consumers false impression that drug has no more safety risk than some other painkillers; Pfizer defends ad
    \\
    \vspace{-1mm}
     \textbf{\textsc{PointGen + cov}}:\\
     public television commercial for pfizer's painkiller celebrex that has attracted attention for both its length and innovative marketing approach is now being criticized for its message. new celebrex ad, which can also be seen on pfizer web site celebrex.com, represents return to television for product after hiatus of more than two years
     \\
     \vspace{-1mm}
	\textbf{\textsc{DeepReinforce}}: \\
	new television commercial for pfizer's painkiller celebrex that has attracted attention for both its length and innovative marketing approach is criticized for its message. \colorbox{red!35}{public citizen, food and drug administration} to ban commercial, charging that it gives consumers false impression that prescription drug has no more safety than some other painkillers.
	\\
	\vspace{-1mm}
    \textbf{\textsc{SENECA}}: \\
    new television commercial for pfizer's painkiller celebrex that has attracted attention for both its length and innovative marketing approach is now being criticized for its message. public citizen, consumer group asks food and drug administration to ban commercial, charging that it gives consumers false impression that prescription drug has no more safety risk than some other painkillers. fda asks pfizer to suspend its television advertising for celebrex
	\\
	\vspace{-1mm}
    \textbf{\textsc{SENECA + $R_{\rm Coh}$}}:\\
    new television commercial for pfizer's painkiller celebrex that has attracted attention for both its length and innovative marketing approach is now also criticized for its message. {\color{blue!75}\underline{public citizen, asked food and drug administration}} to ban commercial, charging that it gives consumers false impression that prescription drug has no more safety risk than some other painkillers. fda asks pfizer to suspend its television advertising for celebrex, and company complied 
    \\
    \hline
	\end{tabular}
	\vspace{2mm}
	\caption{
	Sample summaries for an NYT article. Comparison model contains \colorbox{red!35}{grammatical errors}; our model is {\color{blue!75}\underline{more coherent and with less redundant information}}. Numbers are replaced with ``$0$". 
   }
\label{fig:nyt_sample2}
\end{figure*}

\begin{figure*}[t]
    \fontsize{10}{11}
    \centering
    \setlength{\tabcolsep}{0.8mm}

	\begin{tabular}{|p{160mm}|}
    \hline
    \large{\textbf{\textsc{Article}}}: \\
    \vspace{-3mm}
    many of us may not have the billionaire lifestyle or scientific prowess of marvel's tony stark, but we can now at least get his trademark hi-tech gloves at home. laser expert patrick priebe created a working iron man-style arm and hand that fires beams from the back of the wrist or from the wearer's palm. and in a video, \colorbox{green!45}{the contraption is shown popping balloons and lighting matches} from feet away. scroll down for video. laser expert patrick priebe built a working iron man-style arm and hand that fires beams from the back of the wrist or from the wearer's palm (pictured lighting a match). the gadget was created by wuppertal-based mr priebe, who designs and builds metal laser gadgets to order online. he posts videos to youtube as anselmofanzero. his laser gloves and laser gauntlet/arm are available to order in varying wavelengths. earlier this year, mr priebe built a laser weapon, inspired by the gatling gun. the gun is made from aluminium and shoots rounds of 1.4-watt class 4 blue lasers. the inventor said, although it is only strong enough to pop balloons, it is still too dangerous to be used outside. a video shows the rapid-fire weapon popping 26 balloons mounted on a wall. eight aa batteries power the laser motor and the gun's turret is powered by four ball bearings controlled by a mechanism under the gun. as the barrel spins, the lasers - powered by lithium-ion batteries - are shot out. these range from the blue 405nm-type laser model to one fitted with 445nm, or 650nm. he also offers a ` low power green' version on his website. the gadgets are powered by lithium-ion cells and output ranges between 0.2 w and 1.2 w. prices are not known. (...) \\
    \hline
    \textbf{\textsc{Human}}: \\
    iron man-style arm and hand was built by laser expert patrick priebe. it fires beams from the back of the wrist or from the wearer's palm. in a video, \colorbox{green!45}{the contraption is shown popping balloons and lighting matches}. gadget is powered by lithium-ion cells and can be ordered from mr priebe.
    \\
    \vspace{-1mm}
     \textbf{\textsc{PointGen + cov}}:\\
    laser expert patrick priebe created a working iron man-style arm and hand that fires beams from the back of the wrist or from the wearer 's palm. the gadget was created by wuppertal-based mr priebe, who designs and builds metal laser gadgets to order online. \colorbox{red!35}{he posts videos to youtube as anselmofanzero.}
     \\
     \vspace{-1mm}
	\textbf{\textsc{DeepReinforce}}: \\
    laser expert patrick priebe built a working iron man-style arm. the gadget was created by wuppertal-based mr priebe, who designs and builds metal laser gadgets. patrick priebe created a working iron man-style arm and hand that fires beams from the wearer's palm. laser gloves and laser gauntlet/arm are available to order in varying wavelengths. earlier this year , he built a laser weapon, inspired by the gatling gun. the contraption is shown popping balloons and lighting matches from feet away.
	\\
	\vspace{-1mm}
    \textbf{\textsc{SENECA}}:\\
    laser expert patrick priebe created a working iron man-style arm and hand that fires beams from the back of the wrist or from the wearer's palm. the gadget was created by wuppertal-based mr priebe , who designs and builds metal laser gadgets to order online.
    \\
    \textbf{\textsc{SENECA} + $R_{\rm Coh}$ + $R_{\rm Ref}$ +  $R_{\rm App}$}:\\
    laser expert patrick priebe created a working iron man-style arm and hand that fires beams from the back of the wrist or from the wearer's palm. the gadget was created by wuppertal-based mr priebe , who designs and builds metal laser gadgets to order online. in a video, \colorbox{green!45}{the contraption is shown popping balloons and lighting matches} from feet away.
    \\
    \hline
	\end{tabular}
	\vspace{2mm}
	\caption{
	Sample summaries for an CNN/DM article. Our model overlap most with human summaries with \colorbox{green!45}{information} missed by comparisons. Comparison contains sentence which is \colorbox{red!35}{less coherent and readable}.
   }
\label{fig:cnndm_sample1}
\end{figure*}

\begin{figure*}[t]
    \fontsize{10}{11}
    \centering
    \setlength{\tabcolsep}{0.8mm}

	\begin{tabular}{|p{160mm}|}
    \hline
    \large{\textbf{\textsc{Article}}}: if you're in your early 40s, own a flash car and have started listening to taylor swift and one direction, you are likely to be having a midlife crisis. streaming music service spotify believes it has identified the average age of midlife crises at 42. staff analysed data and found \colorbox{green!45}{users aged around 42 drop their usual playlists} -- which usually contain hits from their youth -- in favour of \colorbox{blue!25}{today's chart toppers from the likes of rihanna and sam smith}. streaming music service spotify believes it has identified the average age of midlife crises at 42 (file picture). spotify and its rivals in the streaming music world are working hard to understand the tastes of their listeners, so they can make better recommendations for them (file picture). `during the teenage years, we embrace music at the top of the charts more than at any other time in our lives. as we grow older, our taste in music diverges sharply from the mainstream up to age 25, and a bit less sharply after that,' explained the company on its insights blog. `we're starting to listen to ``our'' music, not ``the'' music. music taste reaches maturity at age 35. `around age 42, music taste briefly curves back to the popular charts -- a musical midlife crisis and attempt to harken back to our youth , perhaps?' the findings come from a study conducted by ajay kalia, who oversees spotify's `taste profiles' product, which tries to understand people's tastes based on their listening habits. spotify and its rivals in the streaming music world are working hard to understand the tastes of their listeners, so they can make better recommendations for them. (...) \\

    \hline
    \textbf{\textsc{Human}}: \\
    spotify believes it has identified the average age of midlife crises at 42. staff analysed data and found \colorbox{green!45}{users aged 42 drop their usual playlists}. start listening to today 's chart toppers , such as \colorbox{blue!25}{rihanna and sam smith}
    \\
    \vspace{-1mm}
     \textbf{\textsc{PointGen + cov}}:\\
     streaming music service spotify believes it has identified the average age of midlife crises at 42. spotify and its rivals in the streaming music world are working hard to understand the tastes of their listeners.
     \\
     \vspace{-1mm}
	\textbf{\textsc{DeepReinforce}}: \\
    spotify believes the average age of midlife crises at 42. staff analysed data and found users aged 42 drop their usual playlists. spotify believes it has identified the average age of midlife. findings come from a study conducted by ajay kalia.
	\\
	\vspace{-1mm}
    \textbf{\textsc{SENECA}}:
    staff analysed data and found \colorbox{green!45}{users aged around 42 drop their usual playlists} in favour of today's chart toppers from \colorbox{blue!25}{rihanna and sam smith}. streaming music service spotify believes it has identified the average age of midlife crises at 42.
    \\
    \textbf{\textsc{SENECA} + $R_{\rm App}$}:\\
    streaming music service spotify believes it has identified the average age of midlife crises at 42. staff analysed data and found \colorbox{green!45}{users aged around 42 drop their usual playlists} -- in favour of today's chart toppers from \colorbox{blue!25}{rihanna and sam smith}.
    \\
    \hline
	\end{tabular}
	\vspace{2mm}
	\caption{
	Sample summaries for an CNN/DM article. Our models are able to capture \colorbox{green!45}{important information} along with \colorbox{blue!25}{correct entities}. Comparisons suffer from either losing important information or redundancy.
   }
\label{fig:cnndm_sample2}
\end{figure*}
\end{document}